\documentclass[a4paper,11pt]{amsart}
\author{Soumya Kambhampati}
\title{Senior Thesis}

\usepackage{amsmath}
\usepackage{url}

\usepackage{color}
\usepackage{mathtools}

\usepackage{amssymb}
\usepackage{amsthm}

\usepackage{bussproofs}
\usepackage{fullpage}
\usepackage{cite}
\usepackage{latexsym}
\usepackage{listings}
\usepackage{synttree}
\usepackage{textcomp}
\usepackage{verbatim}
\usepackage{enumitem}
\usepackage{booktabs}

\usepackage{hyperref}

\usepackage{graphicx}

\makeatletter

\newcommand*{\reflectmathsymbol}[2]{%
  \reflectbox{$\m@th#1#2$}%
}
\makeatother

\let \Sum \Sigma
\let \i \textit

\theoremstyle{definition}

\usepackage{setspace}
\usepackage[hang]{footmisc}
\usepackage{lipsum}
\makeatletter
\newcommand\ackname{Acknowledgements}
\if@titlepage
   \newenvironment{acknowledgements}{%
       \titlepage
       \null\vfil
       \@beginparpenalty\@lowpenalty
       \begin{center}%
         \bfseries \ackname
         \@endparpenalty\@M
       \end{center}}%
      {\par\vfil\null\endtitlepage}
\else
   \newenvironment{acknowledgements}{%
       \if@twocolumn
         \section*{\abstractname}%
       \else
         \small
         \begin{center}%
           {\bfseries \ackname\vspace{-.5em}\vspace{\z@}}%
         \end{center}%
         \quotation
       \fi}
       {\if@twocolumn\else\endquotation\fi}
\fi
\makeatother

\begin{document}

\begin{titlepage}
    \begin{center}
        \vspace*{1cm}
        
        \textbf{``I ain't tellin' white folks nuthin"}
        
        \vspace{0.5cm}
        A quantitative exploration of the race-related problem of candour in the WPA slave narratives
        
        \vspace{1.5cm}
        
        \textbf{Soumya Chakrabarti Kambhampati} 
        \vfill
        
        A thesis presented in partial fulfilment of the requirements for the degree of\\
        Bachelor of Arts\\
        in \\
        Statistics \& Data Science
        
        \vspace{0.8cm}
                
        Yale University\\
        United States\\
        21 April 2018
        
    \end{center}
\end{titlepage}

\begin{abstract}
	During the Great Depression, the Works Progress Administration interviewed thousands of former slaves about their life experiences. While these interviews are crucial to understanding the ``peculiar institution" from the standpoint of the slave himself, issues relating to bias cloud analyses of these interviews. The problem I investigate is the problem of candour in the WPA slave narratives: it is widely held in the historical community that the strict racial caste system of the Deep South compelled black ex-slaves to tell white interviewers what they thought they wanted to hear. Since no such stiff racial divide existed between the ex-slaves and black interviewers, it stands to reason the topics discussed and sentiments expressed in front of black interviewers differed significantly from those discussed in front of white interviewers. In this work, I attempt to quantitatively characterise this race-related problem of candour. Prior work has either been of an impressionistic, qualitative nature, or utilised exceedingly simple quantitative methodology. In contrast, I use more sophisticated statistical methods: in particular word frequency analysis, sentiment analysis, and comparative topic modelling with Latent Dirichlet Allocation to try and identify differences in the content and sentiment expressed by ex-slaves in front of white interviewers versus black interviewers. While my sentiment analysis methodology was ultimately unsuccessful due to the complexity of the task, my word frequency analysis and comparative topic modelling methods both showed strong evidence that the content expressed in front of white interviewers was different from that of black interviewers. In particular, I found that the ex-slaves spoke much more about unfavourable aspects of slavery like whipping and slave patrollers in front of interviewers of their own race. While these aren't particularly surprising or unknown results, I hope that my more-sophisticated statistical methodology helps improve the robustness of the argument for the existence of this problem of candour in the slave narratives, which some would seek to deny for revisionist purposes. Finally, I found further utility for the results of the comparative topic modelling by using them predict the race of interviews for whom the interviewer was unknown using a simple $k$-nearest neighbours method on the topics discussed in the interview. This method hopefully will allow historians to better utilise the hundreds of interviews for which the race of the interviewer is unknown, which are usually discarded due to their hitherto unascertainable source of bias. 
\end{abstract}

\makeatletter
\@setabstract
\makeatother
\newpage

\begin{acknowledgements}
	There are a few people I would like to thank for their help in crafting my senior thesis. The members of the small S\&DS senior project seminar -- Professor Andrew Barron, Adelaide McNamara, and Keith Woolridge -- all helped immensely with my senior thesis through their suggestions and criticisms during our weekly meetings. I'd like to particularly note Prof. Barron's suggestion that I try to predict the race of the interviewer, a suggestion that directly led to me devising a race prediction method using kNN. I'd like to also thank James Shinn, a History PhD student who initially pointed me to the WPA slave narratives as my TF when I took ``The Rise and Fall of the Atlantic Slave Trade" my Freshman year, sparking an interest in the narratives that culminated in this thesis.
	
	 Above all, I would like to thank my advisor Professor Russell Barbour for his marvellously energetic yet patient guidance of my senior thesis. I will always have fond memories of making the weekly trek to his office at 9 a.m. on Thursdays, and discussing everything from what sorts of significance tests to run to his family's personal experiences during Reconstruction. I will be forever thankful for the critical role Prof. Barbour played in turning what was in effect a back-of-the-napkin idea into a defensible senior thesis.  
	 
\end{acknowledgements}

\newpage

\tableofcontents
\listoffigures
\listoftables
\newpage

\section{Introduction and Prior Work}

Between 1936 and 1938, the Federal Writers' Project, a make-work programme of the Works Progress Administration, interviewed thousands of ageing ex-slaves about their experiences as slaves. These interviews, collectively known as the ``Slave Narrative Collection" are the richest corpus of testimonials of life as a slave in America. As such, they are an incredibly crucial resource in analysing and understanding the ``peculiar institution," as they are one of the only resources which allow us to analyse slavery from the point of view of the slaves themselves. Until almost the 1960s, the testimonials of ex-slaves who had actually lived under the peculiar institution were shamefully ``voiceless" in analyses of slavery \cite{WPA_history}. The Slave Narrative Collection remained effectively untapped until the publication of \i{Roll Jordan Roll} in 1974 by Eugene Genovese \cite{RJR}, an incredibly influential book  considered to be the \i{locus classicus} of discussions of slavery for decades after its publication. The book put forth the thesis that \textit{paternalism} characterised the master-slave relationship. Paternalism, in the words of Genovese, ``brought white and black together and welded them into one people with genuine elements of affections and intimacy" \cite[p.~74]{RJR}. In recent years, Genovese's interpretation has received much push-back for straying too close to the discredited ``plantation myth" narrative of kindly paternalistic masters and grateful slaves. The more modern view is that far from being the centre of the master-slave relationship, paternalism was principally a way to defend the practice of slavery from outside criticism. In the words of Paul Escott, ``Paternalism related more to talk about the plantation than to what actually went on there" \cite[p.~20]{Escott}.  

Where did Genovese go wrong? A plain reading of the slave narratives as authoritative unbiased accounts of what actually went on on the plantation would arguably lead one to similar conclusions to those of Genovese\footnote{Though it is important to note Genovese's interpretation was also coloured by his choice of a Marxist conceptual framework he had previously developed while studying the masters.}, as in many of the interviews ex-slaves professed love for their kindly masters \cite{WPA_limits}. The issue that Genovese seemingly didn't appropriately account for was the issue of candour: did the ex-slaves, when interviewed, actually tell their interviewer what really went on the plantation? Or did they tell their interviewer what they thought they wanted to hear? The vast majority of the interviews were conducted by white people (approximately 1900 of 2358 total interviews), in the 1930s. This context is crucial: these ex-slaves lived in Jim Crow South under an oppressive racial caste system enforced by the terror of the Ku Klux Klan \cite{WPA_limits}. They were old and poor in the midst of the Great Depression, and thus could not afford to alienate their interviewers, who many believed would provide them with some sort of pension \cite{Escott}. Hence it is possible that ex-slaves would refrain from telling the whole truth when it could offend the interviewer, instead giving priority to appeasing their interviewer and telling them what they wanted to hear. Reverend Israel Massie, an ex-slave interviewed in Virginia, put it bluntly: ``I ain't tellin' white folks nuthin' 'cause I'm skeer'd to make enemies"\footnote{This quote is the source for the title of my thesis.} \cite[p.~205]{Weevils}. Martin Jackson, another ex-slave interviewed in Virginia, eloquently described the practice of telling the white interviewer what they wanted to hear: ``Lots of old slaves closes the door before they tell the truth about their days of slavery. When the door is open, they tell how kind their masters was and how rosy it all was" \cite[pp.~219-220]{Weevils}.

This discussion neatly leads to the hypothesis investigated in this thesis: \i{the race of the interviewer significantly impacted what the ex-slave was willing to share about their experiences as a slave, in particular about the varying levels of abuse they suffered}. The strict etiquette of Southern race relations would keep an ex-slave from telling a white interviewer about the cruelty of bondage, but not a black interviewer. So, it stands to reason that the ex-slave would be significantly less reticent when interviewed by a black person. This hypothesis is quite widely held within the historical community, as it is qualitatively easy to see that ``There was more honesty in the all-black interviews and less obeisance to social rituals\footnote{By ``social rituals" Escott is referring to racial etiquette.}" \cite[p.~9]{Escott}; the rub is that it is devilishly difficult to prove. The way it is usually ``proven" in the literature is in the same problematic way most interpretations of the slave narrative collection are proven: the author reads some of the slave narratives, develops some impression from them (no doubt informed by preconceptions), and then writes an article where they use selected quotes from the narratives to buttress their interpretation \cite{Voices}. Call this the ``impressionistic" method of analysis. The issue of course is that the validity of generalisations based on impressionistic or selective evidence is questionable at best. It would be far preferable to support our hypothesis with a systematic study (i.e. quantitative analysis) of the WPA slave narratives \i{in toto}. And that is precisely what I endeavour to do in this thesis.

Systematic statistical studies of the slave narrative collection are few and far between. The book seen to ``most closely approximate a quantitative approach" \cite[p.~379]{Voices} is \i{Slavery Remembered} by Paul Escott \cite{Escott}, published in 1980. Escott, rightly recognising the limitations of an impressionistic approach to the WPA slave narratives, tried for as systematic an approach possible in 1980, reading through all 2358 interviews and hand-classifying them according to eighty-one variables like ``favourability towards master" and ``occupation." He found two pieces of evidence that support my hypothesis. The first was that 72.1\% of ex-slaves responding to white interviewers rated their food quality as good whereas only 46\% of ex-slaves interviewed by blacks did \cite[p.~10]{Escott}. The second, far more powerful bit of evidence was that according to Escott's analysis 26\% of ex-slaves interviewed by white people expressed unfavourable attitudes towards their master, whereas 39\% of those interviewed by black people did \cite[p.~11]{Escott}. In my review of relevant literature, these two pieces of information were ubiquitous in discussions of the race-related candour problem of the WPA slave narratives and are used to ``shore up" the qualitative analysis with some hard numbers. Escott's study, however, suffers from many limitations. While its analysis was impressive in 1980, it seems almost quaint by today's standards, reliant on hand-coding and a microcomputer with less computing power than a smartwatch. The explosion in computing power over the past four decades and the development of advanced statistical tools relating to (in particular) topic modelling provide unexplored avenues to systematically characterise the WPA narratives that I wish to explore in this thesis. Hopefully this analysis can augment Escott's prior quantitative analysis to provide a more compelling argument that the race of the interviewer severely impacted the ex-slaves' candour. 

I would be remiss to not entertain the objections and apprehensions many historians have towards the utilisation of quantitative tools on the slave narrative collection. These concerns derive from the limitations of the slave narratives, in particular the problem of candour I have discussed previously \cite{Voices}. A quantitative tool cannot untangle this and other sources of bias, the argument goes, so it will fall into the same trap that felled Genovese of painting a rosy picture of slavery. This argument holds much merit, I admit, but my investigation side-steps it by narrowing the analysis to this very bias. Indeed, it is possible that by quantifying the bias like I endeavour to do, you could later better apply quantitative methodologies to answer other questions about the slave narratives. 

\section{Methods}
% TODO -- introduction to methods

\subsection{Data Collection and Pre-Processing}
I collected the slave narrative collection manually by hand-scraping the digital transcriptions of the collection from the Project Gutenberg website \cite{PG}.\footnote{I opted to use the Project Gutenberg transcriptions over the ``transcriptions" provided by the Library of Congress because of the readily-apparent superior quality of the Gutenberg transcriptions. In particular, the Gutenberg transcriptions were done manually by a team of dedicated volunteers, whereas the Library of Congress transcriptions were done using optical character recognition with understandably sub-par results.} For each narrative, I recorded the name of the interviewer; if the interviewer was unknown I did not scrape that narrative since it would be of little use. I then matched the names of the interviewers with a list of interviewers and their race determined by Paul Escott \cite{Escott}. While Escott's list of the races of interviewers was the most comprehensive I found, there were still many interviewers whose race was unknown.

Table \ref{interviews_table} reports the number of usable interviews I was able to collect per state. A sizeable proportion of interviews were unusable because there was no record of who the author was, or because the author's race was unknown. All told, I was able to scrape 1,463 usable narratives, of which 286 of the interviews were conducted by black people, and 1,177 were conducted by white people. One thing crucial to note from Table \ref{interviews_table} is that in a few states (like Florida) the majority of the interviews were conducted by black people, and in most states the majority of the interviews were conducted by white people. In some states (like the North Carolina and Ohio), there were \i{no} interviews conducted by black people! This introduces a major potential confounding variable into our analysis, in particular that the difference in candour we are ascribing to the race of the interviewer is actually coming from differences in candour between different states. I try to control for this by running two sets of parallel experiments: one on the entire dataset, and the other only on the dataset of Arkansas interviews. The Arkansas dataset has roughly the same proportion of interviews conducted by black people compared to the full dataset (22\% versus 20\%), and is in fact roughly representative of slavery throughout the South \cite{Escott} since ex-slaves who settled in Arkansas lived through slavery all throughout the South before moving to Arkansas after emancipation. If the results of the analysis on the Arkansas dataset match those on the full dataset, we have some assurance of the robustness of our results. 

\begin{table}[ht]
\centering
\caption{Number of usable interviews conducted by white and black people per state}
\label{interviews_table}
\resizebox{\textwidth}{!}{%
\begin{tabular}{@{}lll@{}}
\toprule
State          & \# of interviews conducted by black people & \# of interviews conducted by white people \\ \midrule
Arkansas       & 161                                        & 574                                        \\
Florida        & 46                                         & 4                                          \\
Georgia        & 38                                         & 123                                        \\
Indiana        & 2                                          & 4                                          \\
Kansas         & 0                                          & 2                                          \\
Kentucky       & 0                                          & 23                                         \\
Maryland       & 17                                         & 38                                         \\
Missouri       & 6                                          & 1                                          \\
Mississippi    & 0                                          & 11                                         \\
North Carolina & 0                                          & 177                                        \\
Ohio           & 0                                          & 67                                         \\
South Carolina & 5                                          & 239                                        \\
Tennessee      & 0                                          & 1                                          \\
Virginia       & 11                                         & 1                                          \\ \bottomrule
\end{tabular}}
\end{table}

I pre-process the narratives by removing excess spacing, removing stop-words, and replacing misspelt words with their corrected counterparts. Many (mostly white) interviewers used deliberate misspelling of words (e.g. ``dey" instead of ``they" or ``massa" instead of ``master") to try and render the dialect of the ex-slaves they were interviewing. Misspelt words could cause problems in my analysis, so I manually corrected all misspelt words that appeared more than 25 times in the dataset. I have made the list of corrections available in the appendix online. Prior to topic modelling, I pre-process further by removing stop-words (frequently seen words considered to have little useful meaning) and lemmatising. Lemmatisation is the process of grouping a word's various inflected forms together (e.g. gather, gathering, and gathering would be treated the same) in an intelligent way so different meanings of a word are treated separately. It (or alternatively word stemming) is considered a standard part of pre-processing prior to topic modelling.

\subsection{Word Frequency and Sentiment Analysis} 
Word frequency and sentiment analysis have the potential to provide powerful evidence supporting the hypothesis. For example, I could use sentiment analysis to see if ex-slaves were more favourable towards their masters (or other aspects of slavery) in front of white interviewers than black interviewers. And even simple word frequencies can speak volumes: for example, if the word ``whip" shows up a lot more in front of black interviewers than white interviewers, this would support our hypothesis that ex-slaves were reticent to speak about abuse in front of white people.

The word frequency analysis consisted of taking simple word frequencies i.e. involved seeing how many times a word of interest shows up in front of white interviewers versus black interviewers, normalised to the length of the document. The words of interest I investigated were:
\begin{enumerate}
	\item \i{whips, whipped, whipping}, to see if ex-slaves were more willing to discuss slave abuse in front of interviewers of their own race
	\item \i{beat, hurt}, for the same reason as (1)
	\item \i{patrollers, patterrollers, pattyrollers, paddyrollers}, to see if ex-slaves were more willing to talk about slave patrols in front of black people. Slave patrols were organised groups of white men who enforced discipline on and policed slaves, in particular runaways and defiant slaves, through whipping, beating, etc. The target words here are different common (mis)spellings of slave patrollers.
	\item \i{rape, raped, raping}, to see if former slaves were more willing to talk about sexual abuse in front of black people. 
	\item \i{bred, breed, breeding}, for the same reason as (4).
	\item \i{happy}, as an indicator to see if former slaves painted a rosier picture of slavery (e.g. ``I was happy as a slave") in front of white people compared to black interviewers.
	\item \i{kkk, klux, klan}, to see if ex-slaves were significantly more likely to discuss the Ku Klux Klan in front of black interviewers. 
\end{enumerate}

Following \cite{Signif_Tests}, I use Welch two-sample t-tests to see if the word frequency differences between the white interviewer and black interviewer datasets are significant \i{if we assume normality}. Unfortunately we cannot assume normality here because the data was collected by nonrandom sampling \cite{Escott}, so the t-tests should only be taken to be ``illustrative." I use the standard $p=.05$ significance level to ascertain ``illustrative" significance.

The sentiment analysis method was very simple. Basically, I found the sentiment of sentences including a target word (e.g. ``master") to measure the overall favourability towards the target word and see if there is a significant difference in favourability for interviews conducted by whites versus black people. I measured the sentiment of a sentence by simply finding all words in the sentence that had a sentiment value (from -1 to 1) in the SentiWordNet sentiment lexicon \cite{SentiWordNet}, and summed the sentiment values up.\footnote{I experimented with a few other lexicons, but found that SentiWordNet was the most comprehensive lexicon with numerical sentiment scoring. My experiments with other lexicons can be found on the online appendix.} I then found the aggregate sentiment of all the sentences including the target word by averaging up the sentiments of the sentences. This is an extremely simplistic methodology with many limitations: it doesn't even account for negation i.e. ``not happy" has a positive sentiment value. I chose it over the state-of-the-art, in particular a recursive neural tensor network  \cite{RNTN_Sentiment}, because it is an extraordinarily explainable model (I can explain precisely why a sentence got a sentiment score) and because the more sophisticated sentiment analysis models I explored had major issues when faced with the misspelt ``dialectical" English of many of the narratives. In particular, the RNTN only operates correctly when all words used are in the Stanford sentiment treebank, which dialectical words like ``marster" are not found in. I then used Welch's 2-sample t-test to illustrate (assuming normality) if there was a significant difference in the overall sentiments.

The sentiment and word frequency analysis methods were implemented in R using the tm \cite{tm} and TidyText \cite{TidyText} packages. The implementation code can be found in the appendix online. 

\subsection{Comparative Topic Modelling}
Comparative topic modelling is a powerful tool that allows us to systematically find and quantify what topics are covered in front of white interviewers versus black interviewers, and to what depth the topics mentioned are covered. As such, it can provide powerful evidence for the existence of the race-related problem of candour. For example, if topics relating to slave abuse are covered more and in more depth in front of black interviewers, this would be strong evidence supporting the hypothesis. I will distinguish between two different methods in my comparative topic modelling analysis: \i{manual} and \i{systematic} comparative topic modelling. But first, I will provide an overview of Latent Dirichlet Allocation \cite{LDA}, the method I use to generate the topic models.

\subsubsection{Overview of Latent Dirichlet Allocation}
Latent Dirichlet Allocation \cite{LDA} is a generative probabilistic model utilised to automatically discover the latent topics present the documents of a corpus and describe precisely the weight of each of these topics within each document. Given a fixed number of topics $K$, LDA represents every document in the corpus as a mixture of these $K$ topics, and each topic as a distribution over words. LDA assumes that the topic distribution has a sparse Dirichlet prior, which can be thought of as encoding the intuition that documents cover a small set of topics and topics frequently use a small set of words. To put it formally, LDA assumes that each document in the corpus was generated by the following procedure. Denote the Poisson distribution with parameter $x$ as $\text{Pois}(x)$, the categorical distribution with parameter $x$ as $\text{Cat}(x)$, and the Dirichlet distribution with parameter $x$ as $\text{Dir}(x)$. For every topic $t$, generate a word distribution $\varphi_t \sim \text{Dir}(\beta)$. Then, generate every document in the corpus as follows:
\begin{enumerate}
	\item Choose the length of the document $N \sim \text{Pois}(\xi)$.
	\item Choose the topic distribution $\theta \sim \text{Dir}(\alpha)$.
	\item Each word $w$ in the document is generated in the following way:
	\begin{enumerate}[label=(\alph*)]
		\item Choose a topic $z \sim \text{Cat}(\theta)$.\footnote{N.B. in the LDA literature the categorical distribution is usually called the multinomial, but since there is only one trial, I opted to more accurately call it the categorical distribution.}
		\item Choose a word $w \sim \text{Cat}(\varphi_z)$ generated by the topic.
	\end{enumerate}
\end{enumerate}

Assuming this generative model, LDA backtracks and tries to infer the topic model, in particular the word distributions of every topic i.e. the probability each word is generated by the topic, and the topic distributions of every document i.e. the probability each document is generated by the topic. There are many ways to perform this inference such as collapsed Gibbs sampling or expectation propagation \cite{LDA}, but the package in R we use, text2vec \cite{text2vec}, utilises WarpLDA \cite{WarpLDA}, which is based on Monte Carlo expectation-maximisation.\footnote{The principal advantage of WarpLDA is its speed; in particular, it's runtime is invariant to the number of topics.} In our analysis we will use the inferred word distribution of each topic to understand what each topic is about, and will use the inferred topic distribution of each document to see how the topics discussed in front of white and black interviewers differed.  

\subsubsection{Systematic Comparative Topic Modelling}
The purpose of systematic comparative topic modelling is to try and see if there is a significant difference between the topic distributions of the white interviewer dataset and the topic distribution of the black interviewer dataset. Intuitively, a significant difference in the topic distributions of the two datasets suggests that there is some difference in the content discussed (which would support our hypothesis). If we don't find a significant difference in the topic distributions of the white and black interviewer datasets, this would suggest that in fact there was no real difference in the topics discussed or in the candour shown, that perhaps the impressionistic qualitative prior work had gotten it wrong. It is illustrative to think of the procedure as a test of significance, where the null hypothesis is that there is no significant difference between the topic distributions of the white interviewer dataset and the black interviewer dataset, and the alternative hypothesis is that there is a significant difference. We test this hypothesis through the following procedure. 

First, we train an LDA model on the entire corpus (both white and black interviewers). This model gives us a topic distribution for each document, a vector $\langle p_{t_1}, ..., p_{t_k} \rangle$ which specifies the probability $p_{t_i}$ that each topic $t_i$ generated the document. Next, we split the corpus into three sets in one of two ways:
\begin{enumerate}
	\item Split the white interviewer dataset into two groups, calling one the train set and the other the validation set. Call the black interviewer dataset the test set.\footnote{These names have other well-established meanings, but they were the best I could come up with.}
	\item Split the black interviewer dataset into two groups, calling one the train set and the other the validation set. Call the white interviewer dataset the test set.
\end{enumerate}
Next, we average the topic distribution vectors of the documents in the training set and those in the validation set. Then, we calculate the Euclidian distance\footnote{The Euclidian distance between two vectors $\vec{a}$ and $\vec{b}$ is defined as $\sqrt{\Sum^n_{i=1}(a_i-b_i)^2}$ where $|\vec{a}| = |\vec{b}| = n$} between the two average topic distribution vectors.\footnote{The use of distance metrics on topic distributions has been done previously in \cite{LDADist}.} Call this the average topic distance between the training and validation sets. This distance tells us the amount of natural fluctuation there is in the topic proportions.  We now calculate the Euclidian distances between the average topic distribution vectors of the training set and the test set. Call this the average topic distance between the training and test sets. Since our null hypothesis is that the white interviewer and black interviewer datasets have no significant difference in topic distribution, if it was true it would stand to reason that the average topic distance between the training and validation sets (two halves of the white interviewer or black interviewer dataset) would not be significantly different from the average topic distance between the training and test sets. And if we found that there was a significant difference between the average topic distance between the training and validation sets and between the training and test sets, this would indicate that there was a significant difference in topic distributions between the white interviewer and black interviewer dataset, and thus that we should reject our null hypothesis. Note that we do not use Welch's t-tests to ascertain ``illustrative" significance since our data is not really tractable to a t-test. This is because we are not dealing with single values but with vectors of values (the topic distributions). Therefore, the way we demonstrate ``significance" illustratively is through this comparison of the Euclidian distance between the training and validation sets and the training and testing sets.\footnote{One way I could have used significance tests would have been to use Welch's 2-sample t-test on the actual Euclidian distances. But I never fully figured out a reasonable methodology for how to do this in time to implement it unfortunately. At any rate, I think the combination of the t-test for illustrative significance in the manual comparative topic modelling and the graphs comparing the average topic distance between train and valid and train and test, as well as the t-SNE model, are sufficient to establish significance.} Finally, to ensure that our result is invariant to the number of topics, we repeat the test for $K = 2,3,...,100$. For $K = 10$, I also will visualise the 10-dimensional topic space in two dimensions using t-distributed stochastic neighbour embedding (t-SNE) \cite{t-SNE}, a nonlinear dimensionality reduction algorithm that is excellent for reducing to 2 or 3 dimensions (compared to, say, principal components analysis). t-SNE models each high-dimensional topic distribution with (in our case) a two-dimensional point that endeavours to preserve the ``closeness" or ``farness" between the high-dimensional vectors (as determined by Euclidian distance). This visualisation will allow us to visually compare and contrast the topic distributions of documents in the white interviewer and black interviewer dataset and see if they appear different.

There are a few things about this method that warrant further discussion. First, I'd like to justify the decision to train the LDA model on the entire dataset of interviews and then compare the topic proportions between the white interviewer dataset and the black interviewer dataset. A seemingly plausible alternative would be to train the LDA model on only one of the datasets, and then ``test" it on the other. An LDA model, after all, can be applied to new documents to classify it according to the already-learned per-topic word distributions. The problem with this methodology can be most explicitly seen if we consider the case that the actual topics (not just their proportions) discussed were different between the white interviewer and black interviewer datasets. Suppose there is some topic, say regarding the beating of slaves by their masters, that doesn't show up at all in the white interviewer dataset but shows up in the black interviewer dataset. And suppose there is some document regarding the beating of slaves in the black interviewer dataset. If we train on the white interviewer dataset, and then test on the black interviewer dataset, we would entirely miss this topic, and since the topic that best classifies the document about the beating of slaves would not be available, the LDA model would classify it with some other, inferior topic, one that maybe doesn't even include the word ``beat" in its word distribution and is in fact about something entirely different. Disturbingly, if the word ``beat" doesn't show up at all in the white interviewer dataset, it wouldn't be in the vocabulary of the topic model and would actually be \i{ignored} when classifying the document about the beating of slaves! Therefore, we see that it is better to train the LDA model on the entire dataset, so we draw our topics (and vocabulary) from all the documents and avoid this sort of problem.

The final aspect of this method that merits further mention is the fact that we split the dataset in one of two ways, either training/validating on the white interviewer dataset and testing on the black interviewer dataset or vice-versa. The purpose of this is to improve the robustness of our results by controlling for differences in the variation of the topic distributions of documents in the white interviewer and black interviewer dataset. If there are differences, this would mean the average topic distance between the training and validation sets would be different for methods (1) and (2), which introduces a potential confounding variable to our results.

\subsubsection{Manual Comparative Topic Modelling}
The purpose of manual comparative topic modelling is to try and explain any differences I found in the topic proportions in the systematic comparative topic modelling. A limitation of the systematic method is that while it can say that there is a difference in the topic distributions between documents of the white interviewer and black interviewer datasets, it doesn't tell us what topics were the cause of this difference and what those topics were about. Manual comparative topic modelling allows us to see which topics in particular were emphasised more in front of white interviewers versus black interviewers.

The methodology for manual comparative topic modelling goes as follows. I trained an LDA model with 10 topics on the entire corpus of interviews (both white and black interviewers). I then identified what each topic was about qualitatively by analysing the word distribution for the topic. I analysed the per-topic word distribution by looking at the words most likely to be generated by the topic (call them top words) for different values of $\lambda \in [0,1]$, the relevance parameter \cite{LDAVis}. Formally, the relevance of a word $w$ to topic $k$ given the weight parameter $\lambda$ is defined as follows: 
\begin{equation}
	r(w,k \mid \lambda) = \lambda \log(\phi_{kw}) + (1 - \lambda) \log(\frac{\phi_{kw}}{p_w}),
\end{equation}
where $p_w$ denotes the marginal probability of $w$ in the corpus and $\phi_{kw}$ denotes the probability of $w$ for topic $k$ \cite{LDAVis}. This formula can be understood as follows: as $\lambda$ goes from 1 to 0, words that are highly generated by other topics are increasingly penalised in the top words list. So $\lambda$ can intuitively be thought of as weighting how ``unique" the top words are to the topic. By using different values of $\lambda$, we can better characterise what each topic is about. In particular, I use $\lambda = 1,.4,.2$ to characterise each topic.

Next, we calculate the average topic distribution of documents in the white interviewer and black interviewer datasets. We define the \i{topic score} to be the vector that results from the element-wise division of the average topic distribution vector of the white interviewer dataset by that of the black interviewer dataset. We compare the average topic distribution vectors and use the topic score to identify which topics have higher proportions in the black interviewer dataset versus the white interviewer dataset. We additionally find the \i{illustrative} significance of the difference in the proportion for each topic using Welch's 2-sample t-test. This gives us a good indication of what sorts of topics ex-slaves chose to emphasise and talk more about in front of black interviewers compared to white interviewers. 

\section{Results and Analysis}

\subsection{Word Frequency Analysis}
The results of the word frequency analysis are reported in Tables \ref{full_wordfreq} and \ref{AR_wordfreq} for the full dataset and the Arkansas dataset respectively. The results are surprisingly elucidatory. 

\begin{table}[ht]
\centering
\caption{Word frequency analysis results on the full dataset}
\label{full_wordfreq}
\resizebox{\textwidth}{!}{%
\begin{tabular}{@{}llll@{}}
\toprule
Word(s)                                                                                                 & \begin{tabular}[c]{@{}l@{}}Average number of times the \\word was spoken in an interview \\with a white interviewer\end{tabular} & \begin{tabular}[c]{@{}l@{}}Average number of times the \\ word was spoken in an interview \\with a black interviewer\end{tabular} & p-value   \\ \midrule
whips, whipped, whipping                                                                          & 0.4078165                                                                                                                   & 1.153846                                                                                                                    & 8.008e-11 \\
beat, hurt                                                                                          & 0.4902294                                                                                                                   & 0.8006993                                                                                                                   & 0.005827  \\
\begin{tabular}[c]{@{}l@{}}patrollers, patterrollers, \\ pattyrollers, paddyrollers\end{tabular} & 0.1860663                                                                                                                   & 0.3321678                                                                                                                   & 0.00675   \\
rape, raped, raping                                                                               & 0                                                                                                                           & 0.003496503                                                                                                                 & 0.3182    \\
bred, breed, breeding                                                                             & 0.03993203                                                                                                                  & 0.1153846                                                                                                                   & 0.004334  \\
happy                                                                                                 & 0.1682243                                                                                                                   & 0.1153846                                                                                                                   & 0.07157   \\
kkk, klux, klan                                                                                   & 0.3738318                                                                                                                   & 0.7167832                                                                                                                   & 0.001038
 \\ \bottomrule
\end{tabular}%
}
\end{table}

\begin{table}[ht]
\centering
\caption{Word frequency analysis results on the Arkansas dataset}
\label{AR_wordfreq}
\resizebox{\textwidth}{!}{%
\begin{tabular}{@{}llll@{}}
\toprule
Word(s)                                                                                                 & \begin{tabular}[c]{@{}l@{}}Average number of times the word \\ was spoken in an interview \\ with a white interviewer\end{tabular} & \begin{tabular}[c]{@{}l@{}}Average number of times the word\\ was spoken in an interview \\ with a black interviewer\end{tabular} & p-value   \\ \midrule
whips, whipped, whipping                                                                          & 0.2874564                                                                                                                          & 2.459627                                                                                                                          & 5.17e-14  \\
beat, hurt                                                                                          & 0.28223                                                                                                                            & 0.9440994                                                                                                                         & 1.134e-05 \\
\begin{tabular}[c]{@{}l@{}}patrollers, patterrollers, \\ pattyrollers, paddyrollers\end{tabular} & 0.1062718                                                                                                                          & 0.3540373                                                                                                                         & 0.0007421 \\
rape, raped, raping                                                                               & 0                                                                                                                                  & 0                                                                                                                                 & n/a       \\
bred, breed, breeding                                                                             & 0.03310105                                                                                                                         & 0.1428571                                                                                                                         & 0.008152  \\
happy                                                                                                 & 0.07142857                                                                                                                         & 0.03726708                                                                                                                        & 0.1195    \\
kkk, klux, klan                                                                                   & 0.5052265                                                                                                                          & 1.198758                                                                                                                          & 6.507e-05 \\ \bottomrule
\end{tabular}%
}
\end{table}

We find that ``whip" (and its inflections ``whipped" and ``whipping") are spoken on average 1.15 times in each interview with a black person but only .41 times in each interview with a white person in the full dataset and 2.46x and .29x respectively in the Arkansas dataset. The $p$-values from the Welch 2-sample t-tests are 8e-11 in the full dataset and 5e-14 in the Arkansas dataset.\footnote{I report the p-values here, but for the other results of the significance tests i.e. the 95\% confidence interval and the degrees of freedom, please refer to the online appendix.} So we can say (assuming normality) that words relating to whipping appear significantly more in front of black interviewer. In addition, when the ex-slave is being interviewed by a black person, the words ``beat" and ``hurt" show up 1.6 times more often ($p$-value of .0058) in the full dataset and 3.4x more often ($p$-value of 1e-05) in the Arkansas dataset. Finally, words relating to slave patrols (in particular ``patrollers", ``patterrollers", ``pattyrollers", and ``paddyrollers") were significantly more likely to be spoken in front of a black interviewer (1.8x with p = 7e-3 in the full dataset and 3.4 with p = 7e-4 in the AR dataset). Taken together, these results suggest that ex-slaves were reticent to discuss being abused by their masters in front of white interviewers but were significantly more candid in front of black interviewers. This provides powerful evidence for our hypothesis that ex-slaves' candour regarding their experiences under slavery was affected by the race of their interviewer. As I discussed in the introduction, this reticence makes sense, given the climate these interviews were conducted: amidst the Great Depression under the rigid racial caste system of the Jim Crow South. 

Interestingly, we also find that this same reticence appears when talking about the Ku Klux Klan; an ex-slave is 1.9 times more likely to speak about Klan in front of a black person than a white person in the full dataset ($p=.001$) and 2.37 times more likely in the Arkansas dataset ($p=$6.5e-05). This makes intuitive sense, since former slaves would understandably be reluctant to tell a white person about the regime of racial terror enacted by the Klan when it is entirely possible the interviewer himself is a Klansman or a KKK sympathiser. We also find rape wasn't really discussed in such words in front of either white or black interviewers; however, ``breed," which is utilised in context to talk about sexual abuse (e.g. breeding slaves together), does appear significantly more often in front of black interviewers. This also supports the hypothesis about candour, as it suggests that ex-slaves were more comfortable talking about sexual abuse in front of members of their own race. The final word frequency test I will discuss is that on ``happy" -- while the word was found 1.5 times more often in front of a white interviewer in the full dataset and 1.9x more often in the Arkansas dataset, the effects were not significant at the standard 5\% level ($p=.07,.12$ for the full and Arkansas datasets respectively). Despite the effect of race on the willingness of the ex-slave to use the word ``happy" not being significant, the fact that the ex-slave was 1.5x and 1.9x times more likely to use the word in front of white people does support our hypothesis that ex-slaves told white people what they thought they wanted to hear, in particular that they were happy and contented, the pernicious ``plantation myth" of slavery. 

Finally, we note that interviews with black interviewers were significantly longer than those conducted by white interviewers. The average length of a interview with a black interviewer was 1020.5 words and the average length of one with a white interviewer was 740.7 words in the full dataset. The p-value of the Welch 2-sample t-test is 4.7e-10. The results I presented above were normalised to the length of each document, so this does not confound our results. However, this result does speak to the candour of ex-slaves: the significantly longer narratives in front of black interviewers does suggest that they were more willing to talk and talk in more length to an interviewer of their own race.\footnote{I should note that this isn't the only possible explanation; another plausible explanation is that white interviewers just didn't write as much of what the ex-slaves wrote down, while black interviewers tried harder to faithfully reproduce precisely what they were told.}

All-in-all, word frequency analysis proved remarkably illustrative, providing large amounts of evidence for our hypothesis. Of particular note, we found that words related to the abuse of slaves appeared significantly more often in front of black interviewers, suggesting that the ex-slaves were reticent to discuss parts of their story that they thought the white interviewer didn't want to hear, but showed significantly more candour when talking to black interviewers.

\subsection{Sentiment Analysis}
To put it simply, the sentiment analysis method unfortunately didn't give me any significant results. This is likely either because of the sentiment lexicon I chose, or more generally the exceeding simplicity of my methodology. As such, I have to summarise my experiments with sentiment analysis to be overall a failure. 
 
 As I explained before, the way I found the sentiment of an ex-slave towards their master was by finding all sentences that included the word ``master," measuring each sentences' sentiment by adding up the sentiment scores of the words from the SentiWordNet lexicon \cite{SentiWordNet}, and then averaging the sentiment of the target sentences written by white interviewers versus black interviewers. The results of this, on the Arkansas dataset, were as follows: the average sentiment score of ``master" on the black interviewer dataset was 0.05956709 and the average sentiment score ``master" on the white interviewer dataset was 0.08635692. Indeed, these results do suggest that the sentiment relating to ``master" was more positive in front of white people as compared to black people. However, the Welch's 2-sample t-test reported a p-value of 0.3582. So the difference we saw was not at all in fact significant. The insignificance is even greater on the full dataset, for which the average sentiment score of ``master" was 0.06814603 and 0.08337082 for the black and white interviewer datasets respectively with a p-value of 0.5274. In a very limited sense our results accorded with the ``ground truth" from Escott's work \cite{Escott}, as favourability towards masters was higher in front of white interviewers (though not significantly so). It is important to note that Escott didn't run any significance tests, so his results may not have been significant either. More importantly he hand-coded passages as either ``favourable" or ``unfavourable" whereas we assign numerical sentiment scores to individual sentences, so we cannot expect our results to precisely accord with his.
 
 My principal apprehension towards the sentiment analysis method comes from exploring the reason sentences were assigned the sentiment they were assigned. This was done on the basis of the sentiment scores provided by the SentiWordNet lexicon, but this lexicon proved to be unsuitable for the task at hand of analysing slave narratives. In particular, the word ``master" had a \i{highly} favourable sentiment value of 0.625! This makes sense in a more general context, since master is used favourably in modern-day speech e.g. ``I mastered the material on the exam," but in the context of analysing slave narrative, this sentiment score makes no sense at all. The results I presented before are with the removal of ``master" as a sentiment word; otherwise the results were worse.\footnote{The results without the removal of ``master" can be found in the online appendix.} So a major issue was that the sentiment lexicon that we used wasn't very well suited for the task. I did experiment with some other readily-available sentiment lexicons, results for which are contained in the online appendix, but none were satisfactory; some, for example, didn't even have a sentiment value for ``whip," an absolutely crucial word in the context of the slave narratives! 
  
\begin{figure}[ht]
\caption{The contribution of top words to the sentiment score for sentences containing ``master" in the Arkansas black interviewer dataset}
\label{sentiment_words_bn}
\includegraphics[width=16cm]{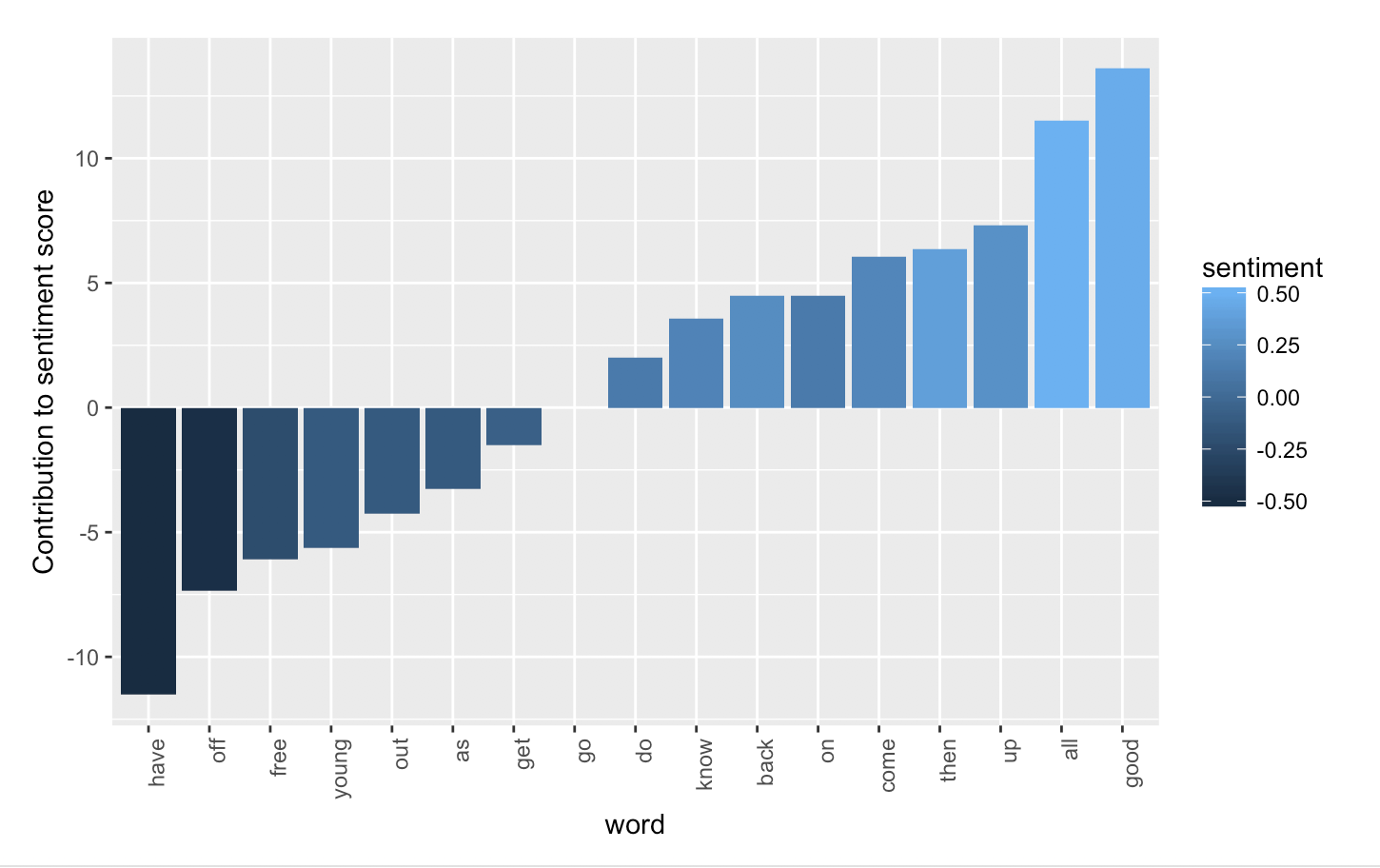}
\end{figure}

\begin{figure}[ht]
\caption{The contribution of top words to the sentiment score for sentences containing ``master" in the Arkansas white interviewer dataset}
\label{sentiment_words_wn}
\includegraphics[width=16cm]{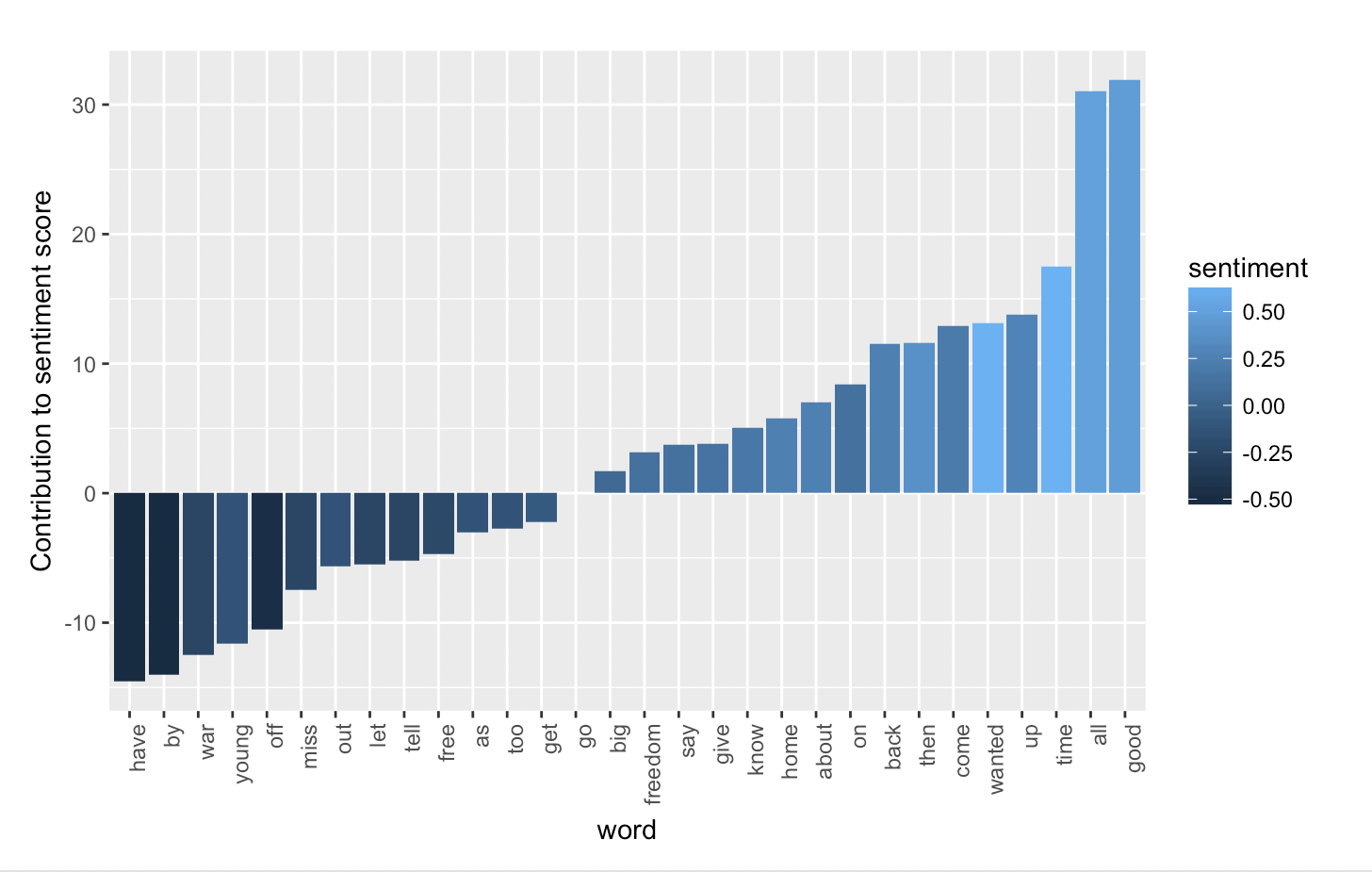}
\end{figure}

 In addition to just having the wrong sentiment value for words, I found that a large proportion of the sentiment value was coming from words that intuitively shouldn't have much weight in calculations of sentiment, at least in the context of analysing slaves narratives. Figures \ref{sentiment_words_bn} and \ref{sentiment_words_wn} provide a visualisation for the contribution of the words to sentiment for the Arkansas black interviewer and white interviewer datasets respectively\footnote{Similar results can be seen for the full dataset, but I thought including them here would be superfluous. They can be viewed in the online appendix.}, calculated by multiplying the word's frequency by its sentiment score according to the SentiWordNet lexicon. While it makes sense that the word ``good" is the largest contributor of positive sentiment, it makes little sense that the word that contributes the most negative sentiment is ``have." After all, intuitively the word ``have" doesn't have any negative sentiment (or really any sentiment) at all. In fact, many of the words that contributed the most to the sentiment words, like ``free" (which has a negative sentiment, which doesn't make any sense in this context at all) or ``young" intuitively ought to not contribute to the sentiment \i{at all}. 
 
 This discussion naturally leads us to consider devising our own domain-specific lexicon free of this issues, or majorly modifying a pre-existing lexicon. I took the first step towards this when I removed ``master" from the lexicon, but I decided not to actually construct my own sentiment lexicon for this task. In addition to time/practicality considerations, I worried that I would be trapped in a chicken-egg problem: to build the lexicon, I would have to determine what sentiment words and sentences in the slave narratives ought to have, but the whole reason we are building the lexicon in the first place is to tell us what sentiment sentences in the slave narratives ought to have! And finally, it might not even make a difference due to the extreme simplicity of our methodology.
 
 I purposefully chose the simplest sentiment analysis method I could find for purposes of explainability, but perhaps a more complex method (with a better lexicon) would perform better. State-of-the-art approaches like the recursive neural tensor network proposed by Socher et al \cite{RNTN_Sentiment} have a much more complex and nuanced understanding of sentiment and human language than merely summing up values according to a lexicon: Socher's RNTN, for example, understands the structure of sentences and uses it crucially to build up sentiment representations of sentences from the sentences' constituent parts.\footnote{A procedure motivated by the famous linguistic principle of semantic compositionality i.e. the meaning of a complex phrase is determined by the meanings of its constituent phrases.} Perhaps Socher's RNTN would be better able to characterise the sentiment of sentences from the WPA slave narratives; such investigations should certainly be carried out in the future.
 
 In sum, due to the insignificance of our results and the major concerns I found with the lexicon I used as well as the overall simplistic nature of the method, I am forced to consider the sentiment analysis portion of my analysis to be an overall failure. 

\subsection{Systematic Comparative Topic Modelling}
\begin{figure}[htp]
\caption{For the Arkansas dataset, the average topic distance between the ``training" half and ``validation" half the black interviewer dataset (the ``baseline") compared to the average topic distance between the ``training" half of the black interviewer dataset and the white interviewer dataset (the ``dist"). }
\label{AR_train_on_bn}
\includegraphics[width=16cm]{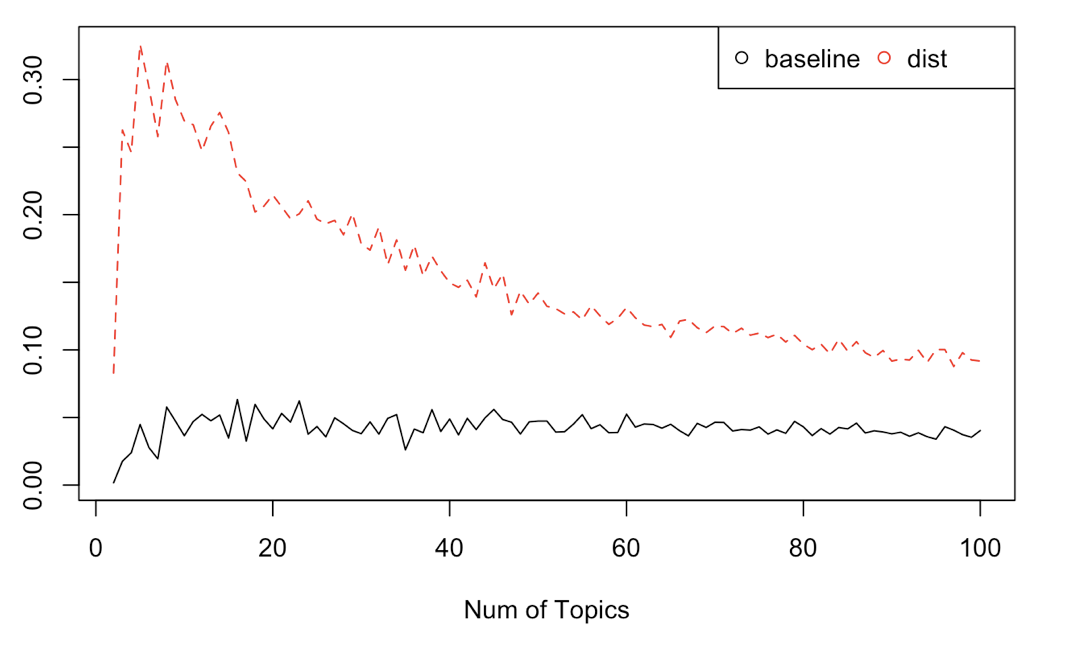}
\end{figure}

\begin{figure}[htp]
\caption{For the Arkansas dataset, the average topic distance between the ``training" half and ``validation" half the black interviewer dataset (the ``baseline") compared to the average topic distance between the ``training" half of the black interviewer dataset and the white interviewer dataset (the ``dist"). }
\label{AR_train_on_wn}
\includegraphics[width=16cm]{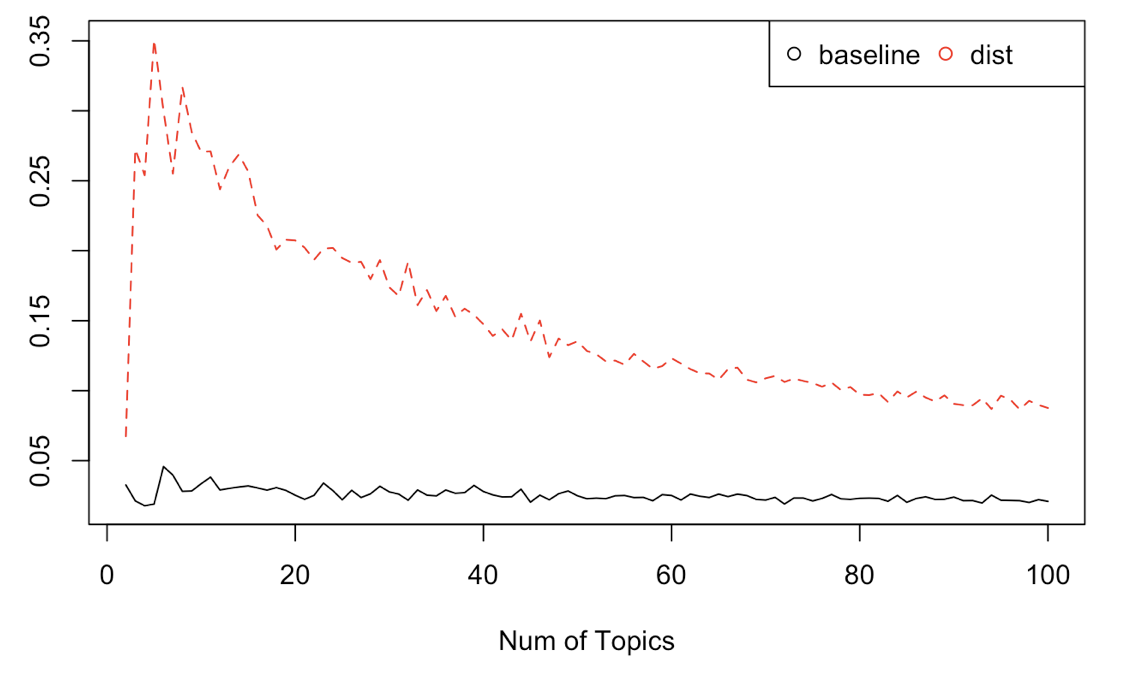}
\end{figure}

\begin{figure}[htp]
\caption{For the full dataset, the average topic distance between the ``training" half and ``validation" half the black interviewer dataset (the ``baseline") compared to the average topic distance between the ``training" half of the black interviewer dataset and the white interviewer dataset (the ``dist"). }
\label{Full_train_on_bn}
\includegraphics[width=16cm]{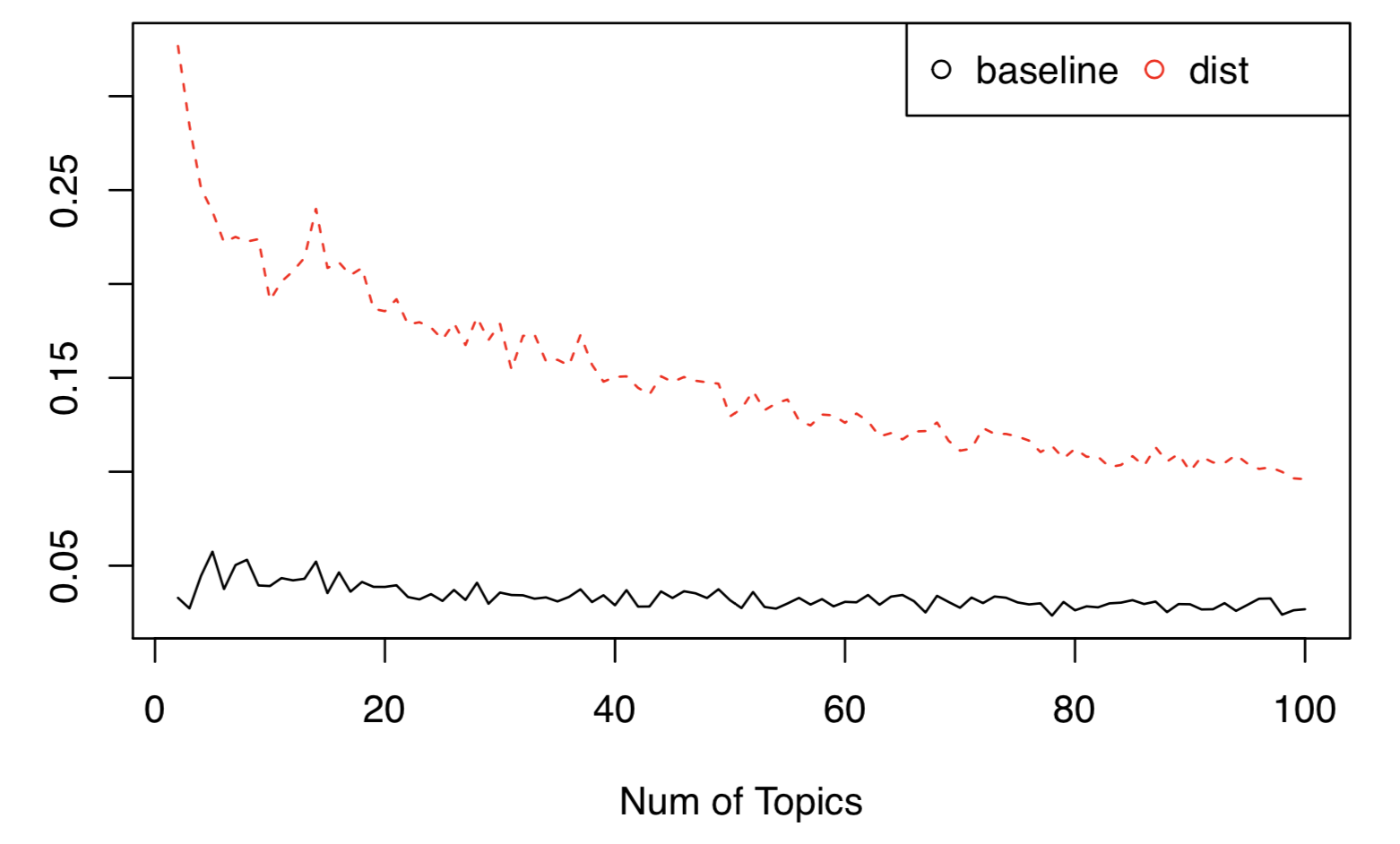}
\end{figure}

\begin{figure}[htp]
\caption{For the full dataset, the average topic distance between the ``training" half and ``validation" half the white interviewer dataset (the ``basline") compared to the average topic distance between the ``training" half of the white interviewer dataset and the black interviewer dataset (the ``dist"). }
\label{Full_train_on_wn}
\includegraphics[width=16cm]{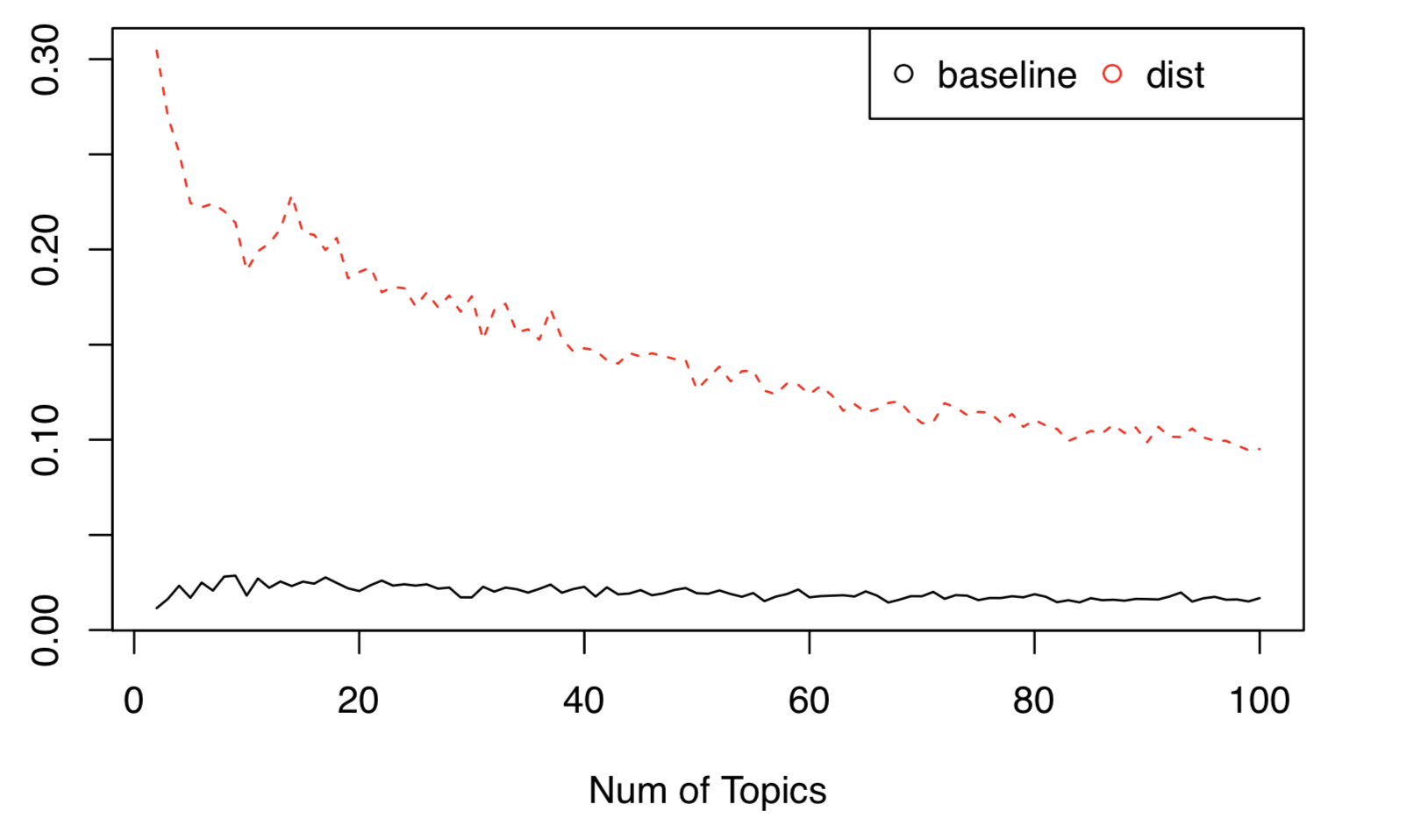}
\end{figure}

The results of the systematic comparative topic modelling are presented in Figures \ref{AR_train_on_bn}, \ref{AR_train_on_wn}, \ref{Full_train_on_bn}, and \ref{Full_train_on_wn}. For simplicity, recall the notation from the Methods section, and use (1) to denote the white interviewer dataset being used for the training and validation sets and the black interviewer dataset being used for the testing set, and (2) to denote the reverse. Figures \ref{AR_train_on_wn} and \ref{AR_train_on_bn} report the cases of (1) and (2) respectively for the Arkansas dataset, and Figures \ref{Full_train_on_wn} and \ref{Full_train_on_bn} report the cases of (1) and (2) respectively for the full dataset. 

We see that the average topic distance between the training and validation sets remains roughly constant as the number of topics increases, while the average topic distance between the training and testing sets decreases. While this behaviour is interesting and certainly worth further exploration, for our purposes it matters little since the average topic distance between the training and testing sets always remains significantly higher than the baseline, even for large numbers of topics. In addition, from my explorations of word distributions for topic models with large numbers of topics (contained in the online appendix), I found the topics to not be very meaningful. In fact qualitatively, I ascertained the 10-topic LDA model provided the most meaningful topics. And the difference between the ``baseline" and the ``dist" was most certainly meaningful at and around 10 topics.

One thing to note is that in Figures \ref{AR_train_on_bn} and \ref{Full_train_on_bn} (the figures reporting the results from splitting the dataset according to method (1)), the ``baseline" is much more jagged than it is in Figures \ref{AR_train_on_wn} and \ref{Full_train_on_wn} (the figures reporting the results from splitting the dataset according to method (2)). This is presumably because the training and validation set for Figures \ref{AR_train_on_bn} and \ref{Full_train_on_bn} are much smaller since they derive from the already-small black interviewer dataset. Another candidate explanation would be that there is a lot more variation in the topic distribution for the black interviewer datasets in Arkansas and for the full corpus compared to that just for the white interviewer dataset. But no matter, the difference still seems to be significant. 

Our results hold for both the Arkansas-only and the full dataset. One difference that can be noticed is that at the left edge of Figures \ref{AR_train_on_bn} and \ref{AR_train_on_wn} (i.e. the Arkansas figures) the average topic distance between the training and testing sets is very low, and steeply increases whereas for Figures \ref{Full_train_on_bn} and \ref{Full_train_on_wn} (i.e. the full dataset figures) it is very high, and steeply reduces. This difference is not substantial, since it happens when there are very few topics, and there are pretty clearly more than \i{two} intuitive topics in the dataset (as I mentioned previously, I found the 10-topic topic model to be the most meaningful).\footnote{Here is a plausible explanation for this behaviour. In the case of the full dataset, LDA made one topic mainly about white interviewer narratives and the other mainly about black interviewer narratives, making the distance between average topic distances quite large. And in the case of the Arkansas dataset, since there are more than 2 topics, LDA made a bad division (say, all dialectical words and all non-dialectical words) for which the average topic distributions between the white and black interviewer datasets was quite small.} 

\begin{figure}[htp]
\caption{t-SNE visualisation of the topic distributions of documents in the Arkansas white (represented by red) and black (represented by blue) interviewer datasets using a 10-topic LDA model}
\label{AR_tSNE}
\includegraphics[width=16cm]{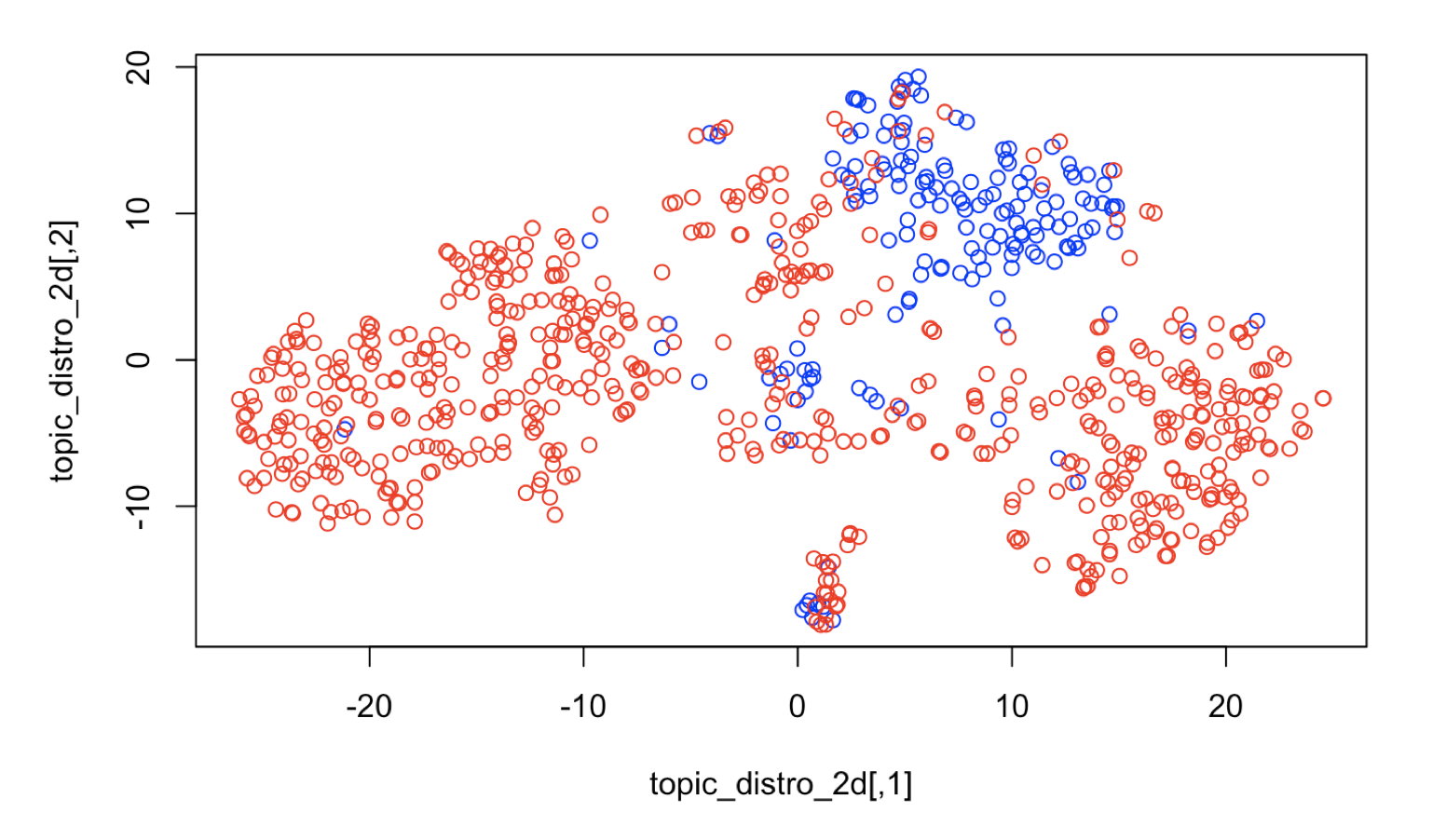}
\end{figure}

\begin{figure}[htp]
\caption{t-SNE visualisation of the topic distributions of documents in the full white (represented by red) and black (represented by blue) interviewer datasets using a 10-topic LDA model}
\label{Full_tSNE}
\includegraphics[width=16cm]{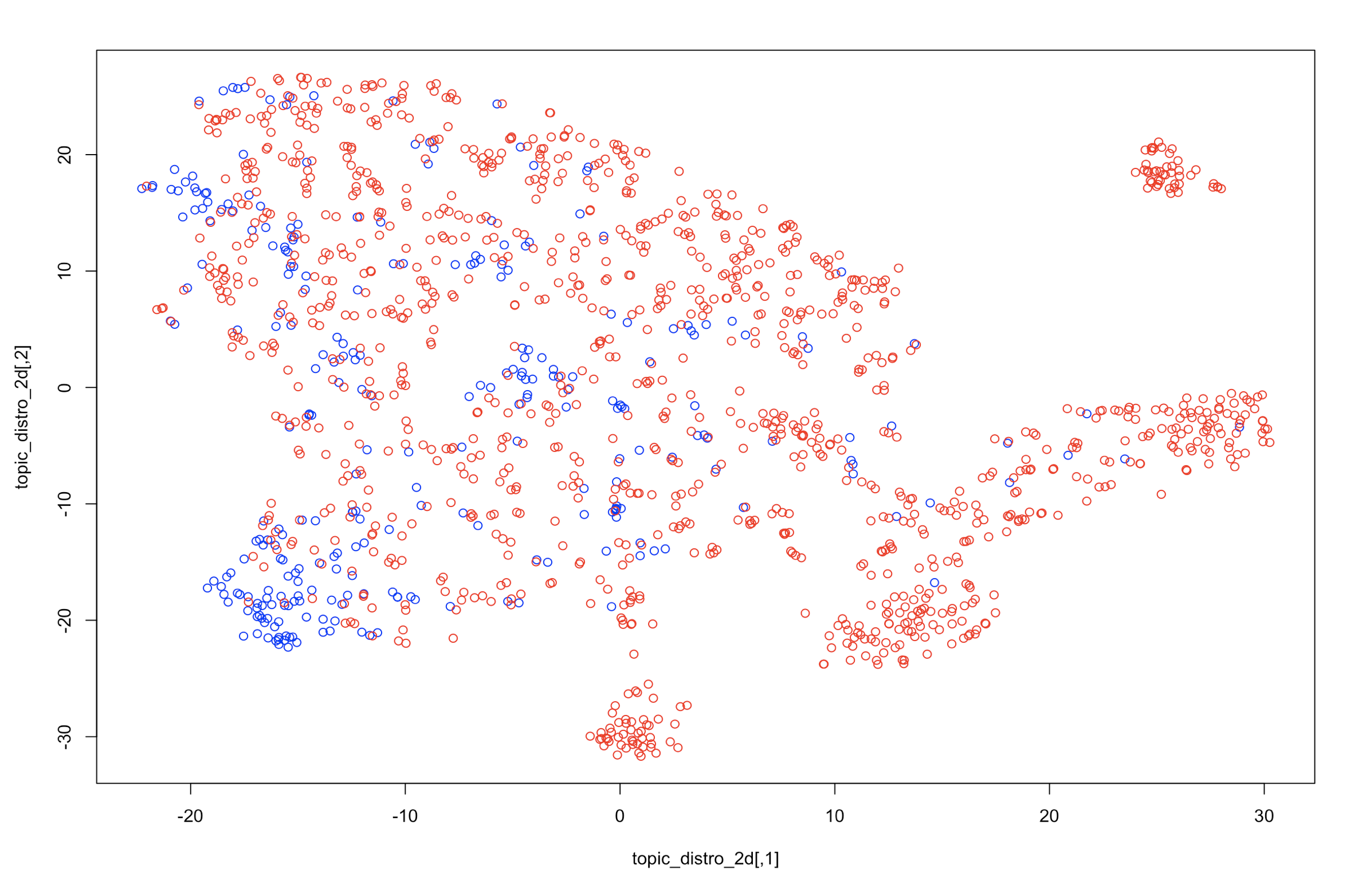}
\end{figure}

Let us now visualise the topic distributions of each documents in the white interviewer and black interviewer datasets using t-SNE as described in the methods section. This way, we might be able to qualitatively ascertain if there is a major difference in the topic distributions. These results are reported in Figures \ref{AR_tSNE} and \ref{Full_tSNE}. Please note that blue dots represent the dimensionality-reduced topic distributions of documents from the black interviewer dataset, and red dots represent the same from the white interviewer dataset.

Qualitatively, we see that the blue dots (representing black interviewer-written documents) in Figure \ref{AR_tSNE} are for the most part quite well separated from the red dots. Figure \ref{Full_tSNE} has less separation, but we can still clearly make out clear clusters of red and blue dots wherein the majority of the dots reside. What these results suggest is that there is indeed a significant difference between the topic distributions of the white interviewer and black interviewer datasets; otherwise the red and blue dots were be far more uniformly interspersed. In addition, the good clustering and differentiation between red and blue we see in Figures \ref{AR_tSNE} and \ref{Full_tSNE} motivate us to try to train a $k$-nearest neighbours model on the topic distribution vectors to predict the race of an interviewer. Being able to predict the race of the interviewer with any sort of accuracy would be hugely important to historians who study the WPA slave narratives. This is because the race of many of the interviewers is unknown, which causes historians to disregard the narratives they wrote because they cannot ascertain this crucial source of bias. 

\begin{figure}[htp]
	\caption{Results of training a k-Nearest Neighbours model using cross-validation on the Arkansas dataset}
	\label{AR_kNN}
	\includegraphics[width=12cm]{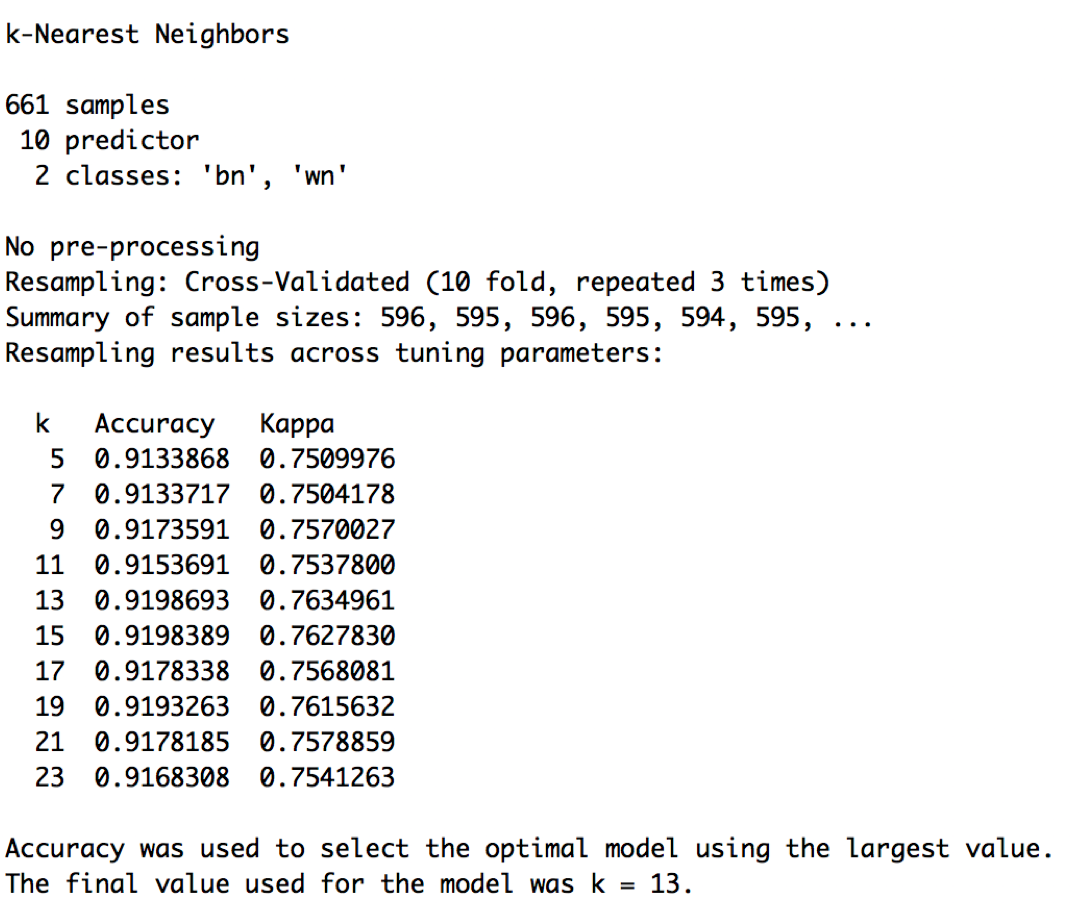}
\end{figure}

$k$-Nearest Neighbours is a simple non-parametric machine learning algorithm that classifies new objects (in this case the topic distribution of a document for which the race of the author is unknown) to whatever class (in this case ``white" or ``black") the majority of the object's $k$ neighbours (in this case the $k$ closest topic distributions by Euclidian distance) belong. You usually use cross-validation to figure out the $k$ that maximises accuracy. The results of my $k$-NN training on the Arkansas dataset is reported in Figure \ref{AR_kNN}. The results on the full dataset can be found on the online appendix.\footnote{The results for the full dataset were not as good, with ~86\% accuracy, but this is still decent.} We were able to get 91.99\% accuracy just using the topic distribution as the only feature! This method can of course be improved by playing with different-sized topic models, feeding in additional features to the model, or using a more sophisticated machine learning method (e.g. a linear model or a SVM). But, importantly for our analysis, we should note that the fact that we had such good performance on just the basis of the topic distribution suggests strongly that there is a significant difference between the topic distributions of documents in the white interviewer dataset and the black interviewer dataset!

All-in-all, these results and the above discussion should convince us that there is a significant and robust difference between the topic distributions of white interviewer dataset and the black interviewer dataset. This finding supports our hypothesis that there was a significant difference in the content of interviews in front of white people and black people. However, we would still like to see what topics are the principal cause of this observed difference. This is the job of the manual comparative topic modelling, the results of which we turn to next.

\subsection{Manual Comparative Topic Modelling}
\begin{figure}[htp]
\caption{For the Arkansas dataset, the top 10 words of the word distribution of each topic for different values of lambda}
\label{AR_topics}
	\hspace*{-1.5cm}                                                           
	\includegraphics[width=18cm]{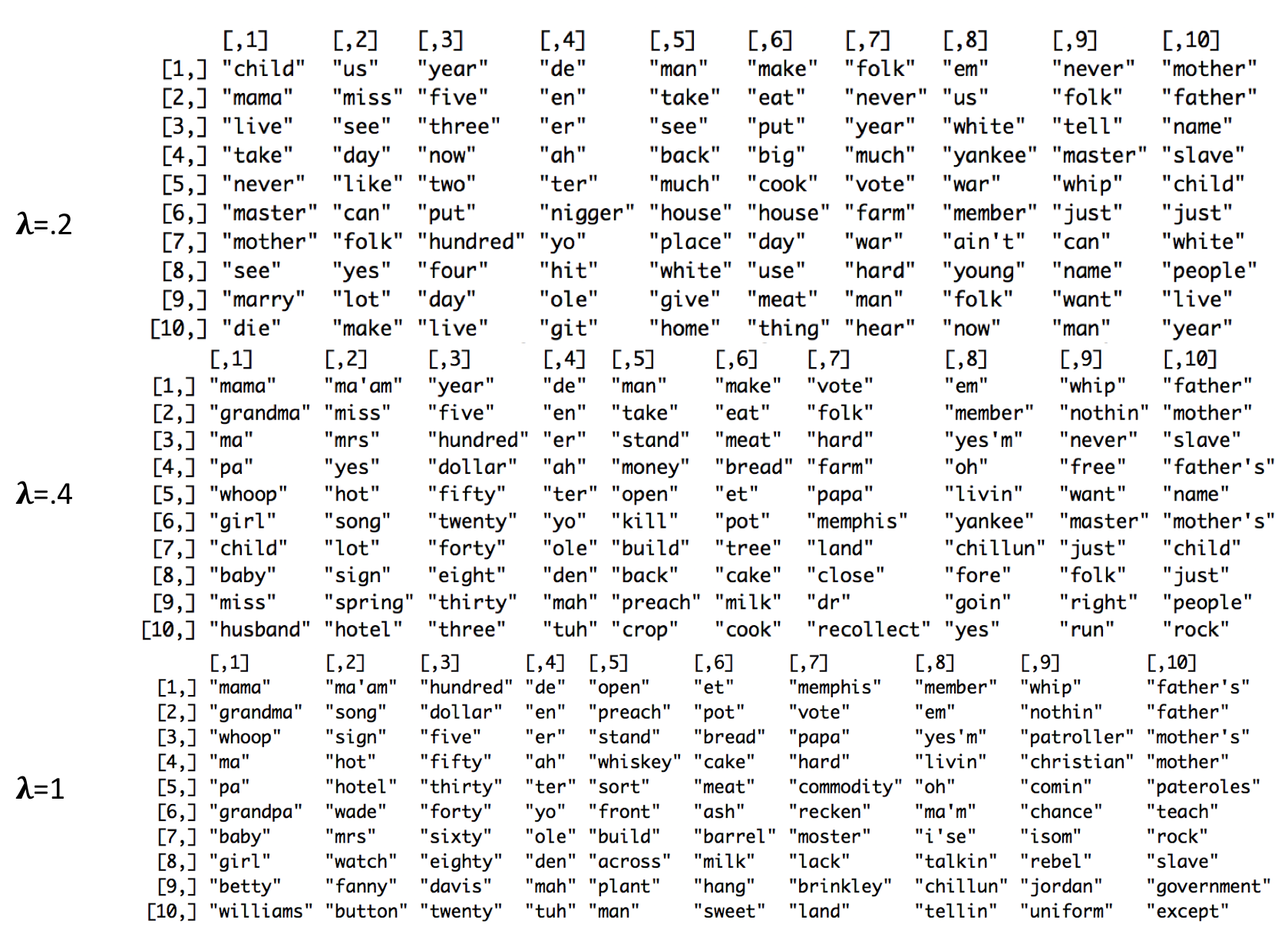}
\end{figure}

I trained a 10-topic LDA model on both the Arkansas dataset and the full dataset. We will first analyse the Arkansas dataset. The word distributions of each topic in the Arkansas dataset is reported in Figure \ref{AR_topics}. From Figure \ref{AR_topics}, we surmise that the topics are about the following:
\begin{itemize}
	\item Topic 9 highly generates the words ``whip", ``patroller", ``run" and ``master." The fact that the topic relates to slave patrollers (who enforced discipline on ``troublesome" slaves and caught runaways), and whipping,  suggests that the topic is the most related to negative descriptions of slavery. This is the topic \textbf{we are most interested in}. 
	\item Topics 5 and 6 seem to be about the more mundane aspects of slavery as they highly generate words like ``bread", ``cake", ``meat", and ``cook". 
	\item Topic 8 highly generates the words ``yankee" and ``war," which suggests the topic is about life during the Civil War.
	\item Topics 1 \& 10 seem to be about family life as they highly generate words like ``mother", ``father", and ``grandma," although Topic 10 seems to be slightly less favourable since it highly generates the word ``pateroles" i.e. slave patrols.
	\item Topic 7 highly generates ``vote" and ``farm," suggesting the topic is about life after slavery.
	\item Topic 4 highly generates a lot of dialectical words (that pre-processing failed to remove), and Topic 3 highly generates a lot of numbers. These topics are meaningless as far as I can tell.
\end{itemize}

\begin{table}[htp]
\centering
\caption{For the Arkansas dataset, differences in the average topic distributions of documents in the white and black interviewer datasets}
\label{AR_topics_table}
\resizebox{\textwidth}{!}{%
\begin{tabular}{@{}lllll@{}}
\toprule
Topic & \begin{tabular}[c]{@{}l@{}}Average probability a \\ document written by a \\ black interviewer \\ generates topic\end{tabular} & \begin{tabular}[c]{@{}l@{}}Average probability a \\ document written by a \\ white interviewer\\ generates topic\end{tabular} & \begin{tabular}[c]{@{}l@{}}Topic score \\ (white interviewer\\ column divided by \\ black interviewer column)\end{tabular} & p-value           \\ \midrule
1     & 0.07095922                                                                                                                     & 0.14354447                                                                                                                    & 2.0229149                                                                                                                  & \textless 2.2e-16 \\
2     & 0.04213053                                                                                                                     & 0.06959749                                                                                                                    & 1.6519489                                                                                                                  & 7.703e-07         \\
3     & 0.11074937                                                                                                                     & 0.07529295                                                                                                                    & 0.6798499                                                                                                                  & 3.257e-08         \\
4     & 0.02844027                                                                                                                     & 0.02242820                                                                                                                    & 0.7886069                                                                                                                  & 0.5314            \\
5     & 0.12299818                                                                                                                     & 0.09935314                                                                                                                    & 0.8077610                                                                                                                  & 0.001533          \\
6     & 0.11327784                                                                                                                     & 0.07999250                                                                                                                    & 0.7061619                                                                                                                  & 0.0002415         \\
7     & 0.03572397                                                                                                                     & 0.18852749                                                                                                                    & 5.2773386                                                                                                                  & \textless 2.2e-16 \\
8     & 0.02907898                                                                                                                     & 0.11560897                                                                                                                    & 3.9756889                                                                                                                  & \textless 2.2e-16 \\
9     & 0.19319256                                                                                                                     & 0.11907039                                                                                                                    & 0.6163301                                                                                                                  & 4.331e-10         \\
10    & 0.25344907                                                                                                                     & 0.08658442                                                                                                                    & 0.3416245                                                                                                                  & 2.2e-16           \\ \bottomrule
\end{tabular}%
}
\end{table}

Now, let's look at the differences in average topic distributions between the white and black interview datasets. This is reported in Table \ref{AR_topics_table}. What we immediately can see here is that the difference between the average topic proportions is significantly different between the white and black interviewer datasets for every topic except topic 4! However, it makes sense that there is no significant difference for topic 4 as it is the topic that primarily contains misspelt dialectical words. This result suggests that the source for the significant difference in topic distances we observed in the systematic comparative topic modelling section is due to significant differences in every meaningful topic. For our purposes, we will restrict our enquiry to the topic we are most interested in, Topic 9. Table \ref{AR_topics_table} tells us that Topic 9, which is about negative descriptions of slavery, is generated on average by an interview conducted by a black person 19.3\% of the time, but only generated 11.9\% of the time when the interview is conducted by a white person. This is to say, there is 64\% more probability mass when the interview is conducted by a black person, and this difference is significant assuming normality (p=4.331e-10). Therefore, we can say that Topic 9, the topic about negative descriptions of slavery, appears significantly more often in front of black interviewers than white interviewers. This is \i{extraordinarily} strong evidence that ex-slaves spoke more candidly about the negative aspects of slavery in front of black interviewers, which is precisely our hypothesis!

\begin{figure}[htp]
\caption{For the full dataset, the top 15 words of the word distribution of each topic for different values of lambda}
\label{Full_topics}
	\hspace*{-1.5cm}                                                           
	\includegraphics[width=18cm]{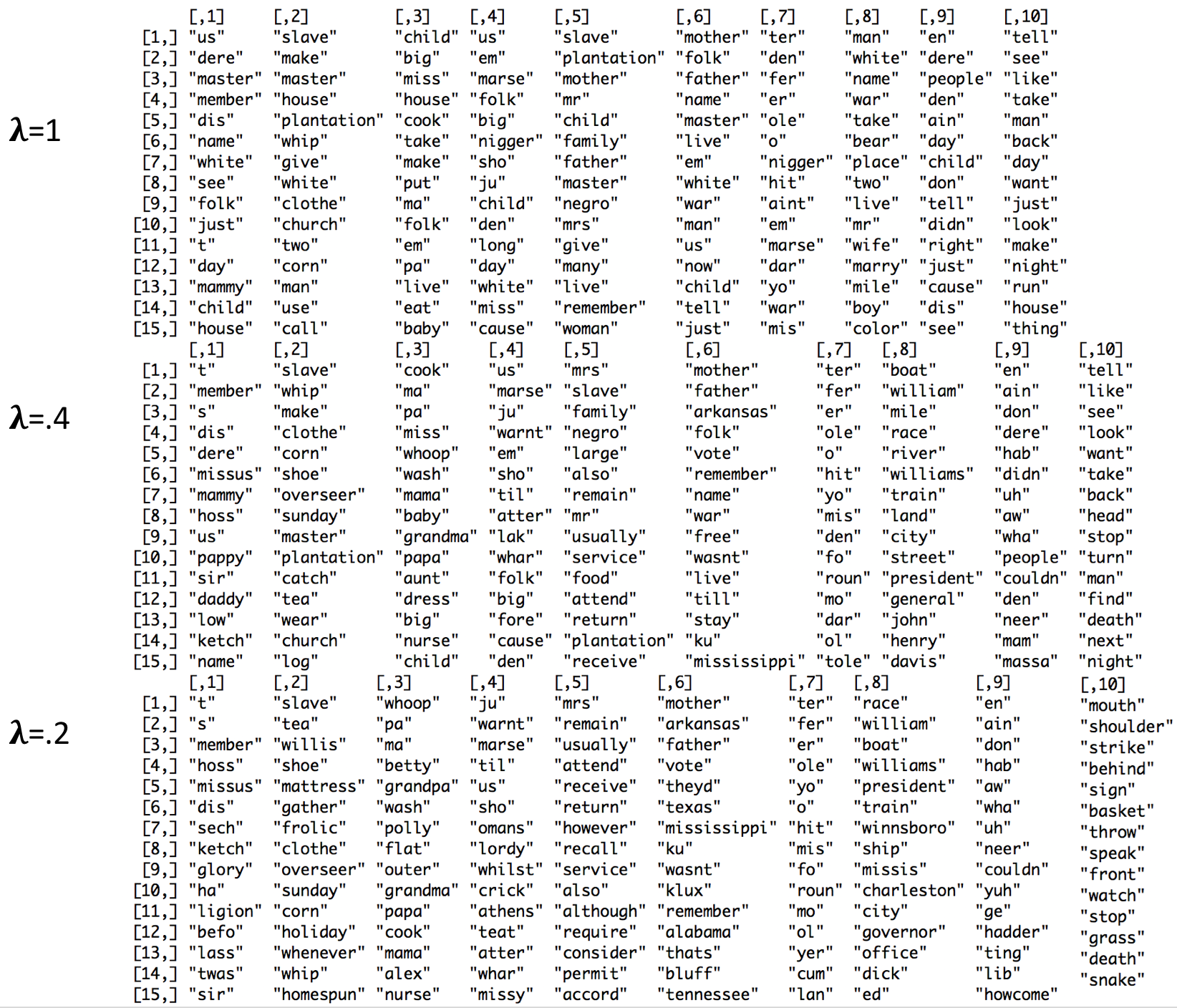}
\end{figure}

Before I go on to the analysis of the full dataset, I want to mention the utility of the manual comparative topic modelling method. I focused on Topic 9 since it was the topic most relevant to my research question, but if you were interested in, say how much the ex-slaves talked about life after slavery, you could compare the topic proportions for topic 7 and find that 19\% of the probability mass of the average topic distribution for white interviewers was devoted to the topic, but only 3.6\% for black interviewers. This suggests that ex-slaves discussed their post-war life significantly more often in front of white interviewers than black interviewers. A possible reason for this effect could be that they were trying to dodge answering questions about their time as a slave by regaling their interviewer with amusing stories of their (post-war) life. At any rate, I hope it is clear that this methodology can be used for a lot of additional analysis of the WPA slave narratives in the future.

\begin{table}[htp]
\centering
\caption{For the full dataset, differences in the average topic distributions of documents in the white and black interviewer datasets}
\label{Full_topics_table}
\resizebox{\textwidth}{!}{%
\begin{tabular}{@{}lllll@{}}
\toprule
Topic & \begin{tabular}[c]{@{}l@{}}Average probability a \\ document written by a \\ black interviewer \\ generates topic\end{tabular} & \begin{tabular}[c]{@{}l@{}}Average probability a \\ document written by a \\ white interviewer\\ generates topic\end{tabular} & \begin{tabular}[c]{@{}l@{}}Topic score \\ (white interviewer\\ column divided by \\ black interviewer column)\end{tabular} & p-value           \\ \midrule
1     & 0.023728048                                                                                                                    & 0.08778015                                                                                                                    & 3.6994256                                                                                                                  & \textless 2.2e-16 \\
2     & 0.141631977                                                                                                                    & 0.07527978                                                                                                                    & 0.5315168                                                                                                                  & \textless 2.2e-16 \\
3     & 0.064484133                                                                                                                    & 0.11093860                                                                                                                    & 1.7204014                                                                                                                  & 2.331e-15         \\
4     & 0.016134428                                                                                                                    & 0.06718095                                                                                                                    & 4.1638260                                                                                                                  & \textless 2.2e-16 \\
5     & 0.185793497                                                                                                                    & 0.05081464                                                                                                                    & 0.2735006                                                                                                                  & \textless 2.2e-16 \\
6     & 0.282763788                                                                                                                    & 0.27029178                                                                                                                    & 0.9558925                                                                                                                  & 0.4368            \\
7     & 0.022084821                                                                                                                    & 0.06825803                                                                                                                    & 3.0907213                                                                                                                  & \textless 2.2e-16 \\
8     & 0.111819696                                                                                                                    & 0.10543440                                                                                                                    & 0.9428965                                                                                                                  & 0.2807            \\
9     & 0.006309451                                                                                                                    & 0.03480884                                                                                                                    & 5.5169371                                                                                                                  & 1.828e-14         \\
10    & 0.145250162                                                                                                                    & 0.12921283                                                                                                                    & 0.8895882                                                                                                                  & 0.01797           \\ \bottomrule
\end{tabular}%
}
\end{table}

Now, to ensure the robustness of our results, we see if they are replicable with the larger full dataset. Figure \ref{Full_topics} reports the word distributions for the 10 topics, and Table \ref{Full_topics_table} reports the differences in average topic distributions between the white and black interviewer datasets. Note that topic 2 highly generates words like ``overseer" (the slave overseer was the white man who maintained discipline in the fields), ``whip," ``catch," ``plantation," and ``master." As such, it seems to be the topic relating to negative descriptions of slavery i.e. the analogue to topic 9 in the Arkansas dataset. To confirm our findings about candour from the Arkansas dataset, we look at the difference in topic proportion for topic 2 between the white and black interviewer datasets using Table \ref{Full_topics_table}. The average probability a document written by a black interviewer generates the topic is .142, but is only .07 if the document is written by a white interviewer! This difference is highly significant assuming normality (p$<$2.2e-16). So, we have confirmed our results from the Arkansas dataset on the full dataset, finding that ex-slaves discussed the negative aspects of slavery (slave patrollers/overseers, whippings, etc.) significantly more often in front of a black interviewer rather than a white interviewer. 

To briefly conclude, we see now that one crucial source of the significant topic distance between the white interviewer and black interviewer datasets is the significant difference in topic proportion for the topic relating to negative descriptions of slavery. This is compelling evidence for our hypothesis that ex-slaves were more willing to speak negatively about their time in slavery in front of a black person than a white person. 

\section{Conclusion and Future Work}
Let us briefly sum-up our important results so far. In our word frequency analysis, we found that words relating to slave abuse appeared significantly more often in front of a black interviewer than a white interviewer. In the systematic comparative topic modelling analysis, we found that the content discussed in front of white interviewers versus black interviewers (as measured by average topic distance) significantly differed. And in the manual comparative topic modelling analysis, we discovered a crucial source for this difference in content discussed: slavery was described negatively significantly more often in front of white people. Our results were effectively the same on the Arkansas and full datasets, suggesting that the effects we observed weren't due the potential confounder of per-state differences but rather due to the race of the interviewer. All in all, these results and the fact that they all coincide provides compelling quantitative evidence that the candour an ex-slave showed was significantly impacted by the race of their interviewer.

In addition to providing strong evidence for the truth of the hypothesis, I devised a method for predicting the race of an interviewer on the basis of their topic distribution using k-Nearest Neighbours. This method really emphasises the importance of the more sophisticated analytical tools I use and their results, since it is not clear if Escott could do such prediction on the basis of his own two statistics. Hopefully, this method allows historians to better understand the bias inherent in the slave narratives for which the race of the interviewer is unknown. In the future, I would like to improve this prediction method by experimenting with different-sized topic models, with different machine learning methods, and with additional features motivated from this study (such as the proportion of words about slave abuse). 

I should also mention the method that didn't work, sentiment analysis. My sentiment analysis methodology was overly-simplistic, and relied on a sentiment lexicon that I found was ill-suited to this specific domain. In the future, I'd like to experiment with modifying the lexicon, and using a more sophisticated sentiment analysis method such as the RNTN model by Socher.

I'd like to end by discussing why I think my work is important. After all, historians for decades have been saying that there are major issues of bias in the WPA slave narratives, in particular due to the lack of candour ex-slaves showed in front of white interviewers (who conducted the vast majority of the interviews). But while they were certainly right about this race-related problem of candour, their reasons for saying so were largely qualitative and impressionistic, and the only systematic quantitative work was very simple and relied on hand-coding. I believe my work fills a major gap on the quantitative side of the argument that there was a race-related problem of candour in the WPA narratives, and thus strengthens the overall argument. This is important, because while historians might accept the bias in the WPA narratives, there are many who deny that the WPA narratives, taken as whole, may be misleading due to this bias. This is because the WPA narratives, accepted uncritically, tell them precisely the story they want to hear: that slave abuse was rare, and that most slaves loved and were grateful to their kindly and paternalistic masters. This plantation-myth of slavery is a pernicious form of historical revisionism that my work hopefully does a small part in fighting back against.

\newpage
\section{Appendix}
The appendix to this paper can be found online at \url{https://tinyurl.com/y9dzx8lu}. The appendix consists of two folders: \i{Analyses}, which contains all of the code and analysis referenced in this paper, and \i{Narratives by State}, which contains all of the narratives I hand-scraped categorised by state and the race of the interviewer. Finally, the appendix includes ``wordstoreplace.csv," a list of all the conversions of dialectical words into their standard English counterparts. Hopefully this appendix is useful both in replicating my results and to assist others who endeavour to do further quantitative research into the WPA slave narratives.

\clearpage
\bibliographystyle{alpha}
\bibliography{references.bib}

\end{document}